\documentclass{article}

\usepackage{microtype}
\usepackage{graphicx}
\usepackage{subfigure}
\usepackage{booktabs} 
\usepackage{enumitem}
\usepackage{multicol}
\usepackage{wrapfig}

\usepackage{algorithm}
\usepackage{algorithmic}
\usepackage{bbm}
\usepackage[table,xcdraw]{xcolor}
\usepackage{acronym}
\usepackage{amssymb}
\usepackage{amsmath}
\usepackage[title]{appendix}
\usepackage{multirow}
\usepackage{dblfloatfix}

\usepackage{listings}
\usepackage{xcolor}

\definecolor{codegreen}{rgb}{0,0.6,0}
\definecolor{codegray}{rgb}{0.5,0.5,0.5}
\definecolor{codepurple}{rgb}{0.58,0,0.82}
\definecolor{backcolour}{rgb}{0.95,0.95,0.92}

\lstdefinestyle{mystyle}{
    backgroundcolor=\color{backcolour},   
    commentstyle=\color{codegreen},
    keywordstyle=\color{magenta},
    numberstyle=\tiny\color{codegray},
    stringstyle=\color{codepurple},
    basicstyle=\ttfamily\footnotesize,
    breakatwhitespace=false,         
    breaklines=true,                 
    captionpos=b,                    
    keepspaces=true,                 
    numbers=left,                    
    numbersep=5pt,                  
    showspaces=false,                
    showstringspaces=false,
    showtabs=false,                  
    tabsize=2
}

\usepackage[font={small}]{caption}

\acrodef{RL}{Reinforcement Learning}
\acrodef{MLSI}{Machine Learning for System Identification}
\acrodef{ADR}{Active Domain Randomization}

\usepackage[final]{corl_2020} 

\title{A User's Guide to Calibrating Robotics Simulators}

\author{
  Bhairav Mehta\\
  Massachusetts Institute of Technology \\ 
  \texttt{bhairavm@mit.edu}
  \And 
  Ankur Handa\\
  NVIDIA \\
  \AND
  Dieter Fox\\
  University of Washington, NVIDIA \\
  \And
  Fabio Ramos \\
  University of Sydney, NVIDIA \\
  
}

\begin{document}
\maketitle


\begin{abstract}
Simulators are a critical component of modern robotics research. Strategies for both perception and decision making can be studied in simulation first before deployed to real world systems, saving on time and costs. Despite significant progress on the development of sim-to-real algorithms, the analysis of different methods is still conducted in an {\em ad-hoc} manner, without a consistent set of tests and metrics for comparison.  
This paper fills this gap and proposes a set of benchmarks and a framework for the study of various algorithms aimed to transfer models and policies learnt in simulation to the real world. We conduct experiments on a wide range of well known simulated environments to characterize and offer insights into the performance of different algorithms. Our analysis can be useful for practitioners working in this area and can help make informed choices about the behavior and main properties of sim-to-real algorithms. We open-source the benchmark, training data, and trained models, which can be found at \url{https://github.com/NVlabs/sim-parameter-estimation}.
\end{abstract}

\keywords{Robotics Simulation, Sim-to-Real, Parameter Estimation, Benchmark} 


\section{Introduction}
Simulators play an important role in a wide range of industries and applications, ranging from drug discovery to aircraft design. In the context of robotics, simulators enable practitioners to test various research ideas, many of which may be too expensive or dangerous to run directly in the real world. As learning-based robotics expands in both interest and application, the role of simulation may become ever more central in driving research progress. However, as the complexity of the task grows, the gap between simulation and real-world becomes increasingly evident. As many recent works have shown, simpler approaches such as uniform domain randomization fail as task difficulty increases, leading to new curriculum and adaptive approaches to minimizing the sim2real gap.

There is recent growing interest in the robot learning community in the field of calibrated simulation and adaptive learning (much of which can be categorized as \textit{machine learning for system identification}). Oftentimes, especially in robotics contexts, it may be reasonable to collect safe demonstrations on hardware using controllers or teleoperation. This line of work, merging the machine learning and system identification communities, is able to incorporate trajectory data into the calibration or policy learning process. Real-world data
can allow for better environment sampling within simulators, more accurate uncertainty estimates for environment parameters, and more robust policy learning.

Calibrating and adaptive simulations hold great promise for robot learning, as a correct set of simulation parameters can help bridge the sim2real gap in a wide range of challenging tasks. When combined with powerful advances in fields like meta-learning or imitation learning, simulation calibration stands to expand the frontier for robot learning applications. With calibrated simulators, learning methods need not trade efficiency for robustness, potentially leading to a future where the simulation itself is embedded into the model-based or closed loop adaptive control loops \cite{Kuindersma:RoboticsToday:BD:2020}.


In this work, we explore current methods in the space of \ac{MLSI}, and push these algorithms to their limits along a variety of axes. We explore failure modes of each algorithm, and present a ``user's guide" on when and where to use each. To present our results cleanly, we introduce the \textbf{Si}mulation \textbf{P}arameter \textbf{E}stimation (\textbf{SIPE}) benchmark, which provides tools to efficiently test and compare past, current, and future algorithms in this space. 


\textbf{Contributions:} Specifically, this work,
\begin{enumerate}
    \item introduces the \textbf{Si}mulation \textbf{P}arameter \textbf{E}stimation (\textbf{SIPE}) benchmark, enabling standardization of benchmarking parameter estimation, system identification, and simulation calibration algorithms;
    \item provides a comprehensive comparison of several current methods across a wide variety of environments, alongside open-source implementations of the algorithms compared, and extensive, varied datasets of trajectories useful in estimation tasks;
    \item compares the usefulness of methods in the accuracy-agnostic task of \textit{sim2sim} transfer.
\end{enumerate}


\section{The SiPE Benchmark}

Robotics benchmarking has seen large improvements in recent years; focus on building simulation environments \cite{openaigym, handenv, tassa2020dmcontrol, James_2020}, robotic trajectory datasets \cite{Cabi_2020, dasari2019robonet}, and teleoperation infrastructure \cite{mandlekar2018roboturk} has proved fruitful for the community, with hundreds of methods annually building upon these tools. Discrepancies between simulators and real world have also been documented \cite{Collins_2020, Menzenbach2019BenchmarkingSA, muratore2019transfer}, yet a lack of standardized tools exist to test such robotic system identification and transfer tasks.

In this section, we introduce the \textbf{S}imulator \textbf{P}arameter \textbf{E}stimation (\textbf{SiPE}) benchmark, a collection of environments, trajectories, and expert policies collected for parameter estimation in robotics settings. As a parameter estimation-focused benchmark, we can abstract away many of the reward design issues that come along with new environment development. In addition, unlike work in reality-simulation discrepancy analysis, the goal is to \textit{automatically} tune parameters of the simulation, rather than manual tuning followed by studying the differences.

\begin{figure}[t!]
    \centering
    \includegraphics[width=\columnwidth]{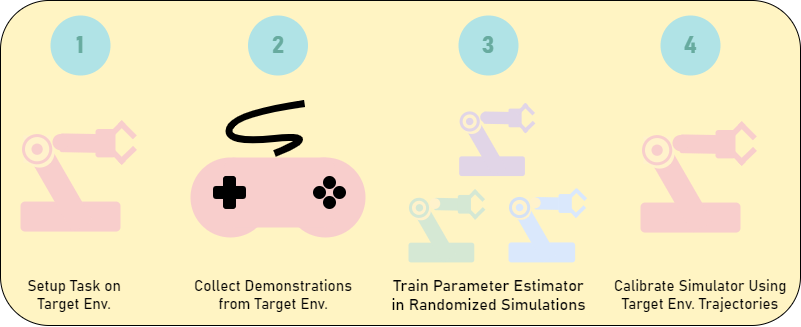}
    \caption{A typical parameter estimation workflow often requires the use of both robotic hardware and simulation, culminating in \textbf{calibration} of the simulator. Better calibration, hopefully, leads to better transfer when using simulators to train control algorithms with reinforcement or imitation learning.}
    \label{fig:parameterestimationdiagram}
\end{figure}

\subsection{Problem Setup}

Parameter estimation in robotics, especially learning based system-identification methods, can often be separated into three distinct stages: 

\begin{enumerate}
    \item Collecting trajectories safely on real robotic hardware.
    \item Training an estimation algorithm to learn how to estimate the parameters from trajectories.
    \item Using real robot trajectories to estimate the parameters of interest and calibrate the simulator. 
\end{enumerate}

As illustrated in Figure \ref{fig:parameterestimationdiagram}, simulators play a significant role in Stage 3. To train system identification algorithms \cite{Jakobi:etal:1997, Abraham:etal:2020, zhou2019environment}, a large diversity of environment setups are needed --- oftentimes, too large a variety to gather the data in the real world. Instead, \textit{domain randomization} \cite{ cad2_rl_sadeghi:etal:2016,domain_randomisation:tobin:etal2017,Peng:etal:2018} allows us to quickly vary simulation parameters of robotic tasks, enabling the large diversity of data to be gathered quickly.

After training, given the reference trajectories from the real world, the normalized values for environment parameters (mass, friction, etc.) that could have generated the trajectories
are inferred. Since the goal of these methods is to calibrate the simulator, we provide each algorithm reference trajectories generated from a test environment, and compare them to trajectories from the environment generated by the estimated parameters. This test environment could be trajectories gathered from the real world robot, or simply a held out set of trajectories from another simulated environment.

Concretely, we define the simulation parameter estimation task as follows: we aim to infer a set of parameters $\theta$, from some real-world trajectories $\tau_{real}$. Given access to a parameterized simulator $S$, we aim to match the real-world trajectory as closely as possible by estimating the simulation parameters $\theta'$ and executing the trajectory in the generated simulator environment $E = S(\theta')$. We use an algorithm $\mathcal{A}$ to estimate the simulation parameters from the trajectories, and calculate a loss $\mathcal{L}$. For notational clarity, we do not denote differences between \textit{training} and \textit{testing} trajectories, which we discuss further in Section \ref{sec:algorithms}. We summarize the general approach in Algorithm \ref{alg:generalizedparamestimation}.

\begin{algorithm}[h!]
    \caption{General Pseudo-Code for Parameter Estimation}
    \label{alg:generalizedparamestimation}
    \begin{algorithmic}[1]
      \STATE {\textbf{Input}: Parameter Estimation algorithm $\mathcal{A}$, Simulator $S$, Number of Parameters $N$, Real robot trajectories $\tau_{real}$}
      \WHILE{not converged}
        \STATE Sample parameters from some prior distribution ($\theta \sim p(\theta)$)
        \STATE Generate simulated environment $E = S(\theta)$
        \STATE Execute trajectory in simulated environment $E$, generating trajectory $\tau_{sim}$
        \STATE Estimate parameters $\theta'$ from trajectory $\tau_{sim}$ and estimation algorithm $\mathcal{A}$
        \STATE Train $\mathcal{A}$ using loss $\mathcal{L}(\tau_{sim}, \tau_{real})$
      \ENDWHILE
      \STATE \textbf{At Inference}
      \STATE Using $\tau_{real}$ and $\mathcal{A}$, calibrate simulator using estimated parameters $\hat\theta$
    \end{algorithmic}
\end{algorithm}

\subsection{Library Design and Implementation}
\label{sec:librarydesign}

\begin{table}[]
\centering
\resizebox{\textwidth}{!}{%
\begin{tabular}{c|ccccc}
\textbf{Environment} & \textbf{Source} & \textbf{Description} & \textbf{Obs. Dim.} & \textbf{Action Dim.} & \textbf{Parameter Dim.} \\ \hline
\textit{FetchPush}         & Plappert \textit{ et al.}\cite{roboticsrfp}  & Push a block to goal & 31 & 4 & 2 \\ \hline
\textit{FetchSlide}        & Plappert \textit{ et al.}\cite{roboticsrfp} & Slide a puck to goal &  31 & 4 & 2 \\ \hline
\textit{FetchPick\&Place}  & Plappert \textit{ et al.}\cite{roboticsrfp} & Pick\&Place block onto other &  31 & 4 & 2 \\ \hline
\textit{Relocate}          & Rajeswaran\textit{ et al.}\cite{handenv}  & Move ball with hand to location & 39 & 28 & 2 \\ \hline
\textit{Door}              & Rajeswaran\textit{ et al.}\cite{handenv} & Open door with hand & 39 & 28 & 7 \\ \hline
\textit{HalfCheetahLinks}  & Brockman\textit{ et al.}\cite{openaigym} & Locomote until termination & 17 & 6 & 5 \\ \hline
\textit{HalfCheetahJoints} & Brockman\textit{ et al.}\cite{openaigym} & Locomote until termination & 17 & 6 & 15 \\ \hline
\textit{HumanoidLinks}     & Brockman\textit{ et al.}\cite{openaigym} & Locomote until termination & 376 & 17 & 17 \\ \hline
\textit{HumanoidJoints}    & Brockman\textit{ et al.}\cite{openaigym} & Locomote until termination & 376 & 17 & 51 \\ \hline
\end{tabular}%
}
\caption{List of environments used in this work to benchmark.}
\label{tab:environments}
\vspace{-20pt}
\end{table}

In order to standardize evaluation of these algorithms, we have compiled a representative set of environments, which range from simple sanity checks to difficult high-dimensional parameter estimation problems. A complete list of characterized environments can be seen in Table \ref{tab:environments} and visualized in Figure \ref{fig:allenvironments}.

Broadly, we consider the following types of tasks:

\begin{enumerate}
    \item \textbf{Object Centric, Simple} - The \textit{Fetch} robotics suite of tasks, from \cite{roboticsrfp}, is a low-dimensional task with only cubes. The parameters estimated here are related to the cube's friction and mass, and serve as sanity checks.
    \item \textbf{Manipulation} - The \textit{Adroit} robotics tasks \cite{handenv} involve a high-dimensional observation and action space, but is a low-dimensional parameter estimation task involving properties of the object the hand interacts with.
    \item \textbf{Locomotion, Simple} - The standard reinforcement learning benchmarks are split into two types of parameter estimation tasks. The simple variants require estimation of link lengths.
    \item \textbf{Locomotion, Difficult} - The more difficult variants of locomotion estimation tasks involve estimating three parameters for each joint. 
\end{enumerate}

For each environment, SiPE provides wide variety of tools and datasets to explore different parameter estimation problem settings. Each SiPE environment comes with:

\begin{itemize}
    \item \textit{Pretrained Agents}: Agents are trained with DDPG \cite{Lillicrap:etal:2015} (training is elaborated on in Section \ref{sec:results}) and include both a optimal and suboptimal ($50\%$ of training completed) agent that can be used to generate trajectories. In this analysis, we use the fully-trained agent to collect trajectories (Stage 2 in Figure \ref{fig:parameterestimationdiagram}).
    \item \textit{Reference, In-Distribution, and Out-Of-Distribution Variants}: To test generalization, we provide environment variants (\textit{i.e.} different parameter settings) that are not seen during training of the estimation algorithm: one within the parameter estimation range, and one outside.
    \item \textit{Reference, Validation, and Test Trajectories from each environment}: To ensure fair comparison and to mitigate overfitting, each proposed SiPE environment comes with three sets of unique trajectories from each proposed environment. In our work, the reference trajectories are used during training, whereas validation and test trajectories are only used to tune hyper-parameters or evaluate performance. 
    \item \textit{Comprehensive Description of Environments and Parameters}: Each environment has been documented to ensure the experimenter knows the effects of each parameter, including how it interacts with the others. 
\end{itemize}

The reference trajectories are generated from a ``default" environment, and each algorithm predicts a (normalized) parameter estimate $\theta \in [0, 1]^N$, where $N$ is the dimensionality of the parameter estimation space. As real robots were not accessible for many of the experiments, we use this reference as the \textbf{target} environment. 

To compare performance across environments and algorithms, we introduce the SiPE plot, inspired by \texttt{bsuite} \cite{Osband:etal:2019} and implemented with Matplotlib \cite{Hunter:2007}, shown in Figure \ref{fig:adr-raw}. Each algorithm is run against all of the benchmark environments in Table \ref{tab:environments}, and the score for each radial point represents the accuracy on that particular environment. Scores are calculated by averaging across all of the parameters. 

However, as many parameter estimation tasks are high dimensional, it can be difficult to compare estimators (especially as compared to reinforcement learning, where cumulative rewards can be compared across agents); SiPE comes with extra comparison modes (\texttt{mean}, \texttt{max}, \texttt{min}, seen in Figure \ref{fig:max-sipe} and \ref{fig:min-sipe} in the Appendix), which, when compared in conjunction, can provide real insight to estimator strengths and weaknesses. Similar to \texttt{bsuite}, SiPE ships with quality-of-life improvements, such as in-depth parameter estimation error plots, \LaTeX report generation, and more.

\subsection{Algorithms}
\label{sec:algorithms}

In order to get a clear understanding of current approaches, we benchmark a few representative methods, listed in Table \ref{tab:estimators}, and compare them within the SiPE tasks described in the previous section.
As the interest in this direction has grown, so have the number of approaches to do calibration or transfer. It is therefore imperative to understand the strengths and weaknesses of these approaches, which we compare across a wide variety of tasks and ablation studies.
While this work may not fully represent analysis of all estimation methods both in use and development, we believe that even this sampling (along with Section \ref{sec:ablations}'s ablations) constitutes a good representation of today's \ac{MLSI} landscape.

\begin{table}[]
\centering
\resizebox{\textwidth}{!}{%
\begin{tabular}{c|cccc}
\textbf{Estimator} & \textbf{Description} & \textbf{Optimization Goal} & \textbf{Particle Based?} & \textbf{Learned Rewards?} \\ \hline
\textit{Regression} & Linear Regression (Baseline) & Cost Minimization & No  & No   \\ \hline
\textit{BayesOpt}   & Bayesian Optimization over parameter space & Cost Minimization  & No  & No \\ \hline
\textit{MAML}       & Regression from \cite{maml} & Fast Adaptation & No  & No   \\ \hline
\textit{SimOpt}     & Particle-based REPS & Cost Minimization & Yes & Both \\ \hline
\textit{ADR}        & Stein Variational Policy Gradient & Cost Minimization, Diversity & Yes & Both \\ \hline
\textit{BayesSim}   & Posterior Estimation of Parameters & Accurate Posterior & No & No \\ \hline
\end{tabular} 
}
\caption{An overview of the estimators benchmarked in this paper.}
\label{tab:estimators}
\vspace{-20pt}
\end{table}

Briefly, we describe the five algorithms and their variants tested in this survey. Some of these algorithms have been adapted to the particular \ac{MLSI} setting we use here, but each represents a reasonable method to do parameter estimation. A full discussion on design choices and hyperparameter search can be found in Appendix \ref{app:hyperparameters}.

\begin{enumerate}
    \item \textbf{Regression}: As a baseline, we use a simple, linear regression approach, trained with stochastic gradient descent. Given trajectories, the goal is to regress the environment parameters using the MSE between trajectories generated by executing the same actions in both the reference (target) environment and the estimated parameter environment. We use the iterative, gradient descent based version, rather than the closed form expression.
    
    \item \textbf{Bayesian Optimization (BayesOpt)}: A popular system-identification tool, we implement a Bayesian Optimization approach, which is optimized using the upper confidence bound acquisition function. As a substitute for utility, we optimized for the lowest MSE between the trajectories generated by executing the same actions in both the reference (target) environment and the estimated parameter environment.
    
    \item \textbf{Model-Agnostic Meta-Learning (MAML)}: Using the regression task setup from \cite{maml}, we adapt the algorithm to the parameter estimation setting. MAML, as a gradient-based meta-learning algorithm, trains using a \textit{meta-training} set, which consists of randomized simulation parameter settings and trajectories rolled out in those environments. At test time, the trained MAML model is given a trajectory from the target environment, and uses a set amount of gradient updates to regress the correct environment parameters. While there have been large numbers of follow up work to improve on the base algorithm \cite{nichol2018firstorder, finn2019online}, we use the original variant as a comparison.
    
    \item \textbf{SimOpt}: SimOpt \cite{simopt} evolves a distribution of particles --- which control the simulation parameters proposed --- towards some ground truth distribution that is inferred from the real-world trajectories provided. SimOpt uses Relative Entropy Policy Search \cite{peters2010reps} to evolve the parameters, using a trajectory discrepancy loss (between real and simulation trajectories) and a trust-region to keep the optimization constrained. 
    
    \item \textbf{Active Domain Randomization}: Active Domain Randomization (ADR) \cite{adr} uses particles, trained with learned rewards based on trajectory discrepancies (similar to the discriminator approach from \cite{eysenbach2018diversity}), to estimate parameter settings. \cite{mehta2020adrplus} proposes a recent improvement that builds upon ADR that aims to infer the parameter settings by ``fitting" to real world trajectories, by flipping the sign of the reward signal. This forces particles to generate parameter settings that generate trajectories that are indistinguishable from the reference (as compared to the original approach that attempted to maximize discrepancies).
\end{enumerate}

As a comparison to the class of parameter estimation algorithms that generate posteriors, we compare the previous algorithms against \textbf{BayesSim} \cite{bayessim}. BayesSim learns a posterior over simulation parameters given real-world trajectories. The algorithm trains a Mixture Density Network \cite{Bishop1994MixtureDN}, allowing for efficient sampling of the posterior from the underlying Mixture of Gaussians distribution. BayesSim, and similar Bayesian Inference algorithms, are built on the assumption that having uncertainty over parameters can be critical in preventing overfitting to simulation, especially as many unique combinations of simulator settings can generate the same trajectory. We introduce a BayesSim comparison in Section \ref{sec:domainrand}, but do not compare it against the previous algorithms in parameter fitting settings, due to the difficulty of comparing Bayesian inference algorithms and their outputs (posteriors over parameters) against the point estimates of parameters found by the other surveyed algorithms.

For approaches that require the use of learned rewards, we train a discriminator to differentiate among types of trajectories, where indistinguishable trajectory sources (i.e the discriminator cannot tell if the trajectory came from the reference environment, or the estimated one) are rewarded higher. For all other estimators, we use a MSE loss between trajectories, executing the same actions from the same starting state in both the reference and proposed environment. 

To ensure fair comparison, and to test the generality of each algorithm, we use \textit{FetchPush-v1} and \textit{Relocate-v0} to hyperparameter tune the approaches, and use the best performing set to benchmark on the rest of the environments. However, as noted in Section \ref{sec:librarydesign}, \textit{all} SiPE environments come with validation datasets, which could be used in more practical, results-driven settings.

\begin{figure}
    \centering
    \begin{lstlisting}[language=Python]
        import sipe 
        import gym
        
        target_env = gym.make('target-env-v1')
        trajectories = sipe.demonstrations(target_env, type='expert')
        
        randomizable_env = sipe.make('randomized-env-v1')
        estimator = get_estimator(randomizable_env, estimator_type)
        for iteration in range(n_max_iterations):
            estimator.update_parameter_estimate(env)
            
        final_estimates = estimator.get_parameter_estimate(env)
        sipe.show_results(final_estimates)
        
    \end{lstlisting}
    \caption{A code snippet that imports, runs, and plots results using the SiPE library.}
    \label{fig:codeexample}
\end{figure}

\section{Results}
\label{sec:results}
\label{sec:ablations}

\begin{figure}
    \centering
    \includegraphics[width=1.0\columnwidth]{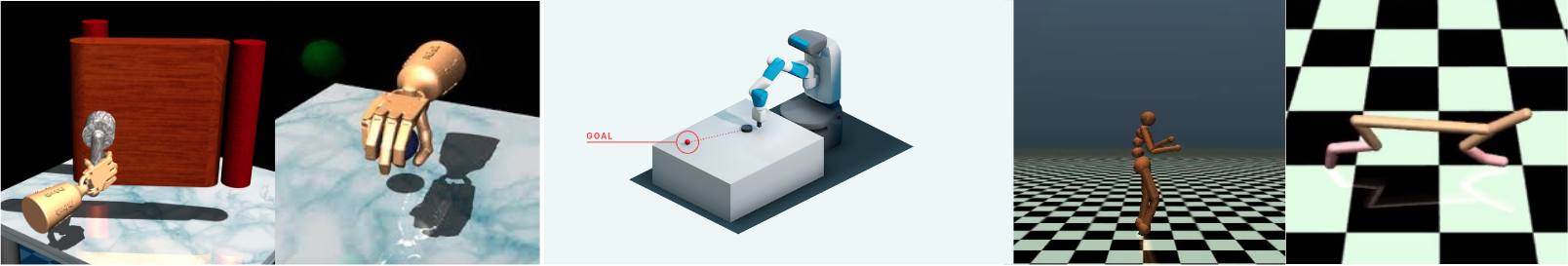}
    \caption{The base environments tested: two complex hand environments from \cite{handenv}, three, lower-dimensional Fetch Robotics environments from \cite{roboticsrfp}, and two locomotion environments from \cite{openaigym}. The locomotion environments are edited to have two variants: a simpler parameter estimation task for link masses, and a more difficult, higher-dimensional joint properties estimation task.}
    \label{fig:allenvironments}
\end{figure}

We benchmark all five algorithms on the environments shown in Figure \ref{fig:allenvironments}, and present the raw results below in Figures \ref{fig:rawresults} with \textbf{results averaged over three trials for each experiment}. In these SiPE plots, only a single estimator is shown per plot, with \texttt{mean}, \texttt{min}, and \texttt{max} accuracies; for example, the \textit{min} plot will take the minimum accuracy across all $N$ parameters estimated. This allows us to conclude general trends of estimators, before exploring and comparing estimators against one another in Sections \ref{sec:policylearning} to \ref{sec:domainrand} and in Appendix A. \textbf{The rest of this section contains our most general analysis of results, and can be used as a self-contained section with which to get started with simulation calibration and \ac{MLSI}.\footnote{For brevity, our results regarding Bayesian Optimization and Linear Regression can be found in the Appendix.}}

By comparing the \textit{spread} --- variance in the results from $\min$ and $\max$ of parameters --- of each estimator in Figure \ref{fig:rawresults}, we can get an understanding of how these estimators behave, even in high-dimensional parameter estimation tasks. We experiment with the following settings:

\begin{itemize}
    \item Decoupling policy learning from parameter estimation by directly regressing for parameters using trajectories from the target environment.
    \item Parameter estimation and policy learning in the loop with hand-tuned rewards.
    \item Parameter estimation and policy learning in the loop but with learned rewards with a discriminator.
    \item Assessing the generalisation of the learned parameters on a suite of test environments.
\end{itemize}

In summary, our conclusions are that particle based methods like ADR and SimOpt tend to do well on average but with larger spread for direct parameter regression. We also notice that simultaneously learning a policy alongside parameter estimation stabilizes all algorithms, despite the fact that performance may drop in certain tested environments. We find no general improvement from learning rewards (when compared to using the MSE error between states).

\subsection{Direct Parameter Estimation}
\label{sec:dpe}

\begin{figure}[h]
  \centering
  \subfigure[]{%
    \includegraphics[width=.3\columnwidth]{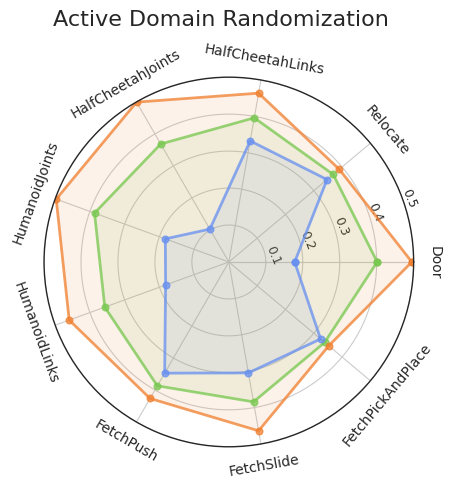}\label{fig:adr-raw} 
  } 
  \subfigure[]{%
    \includegraphics[width=.3\columnwidth]{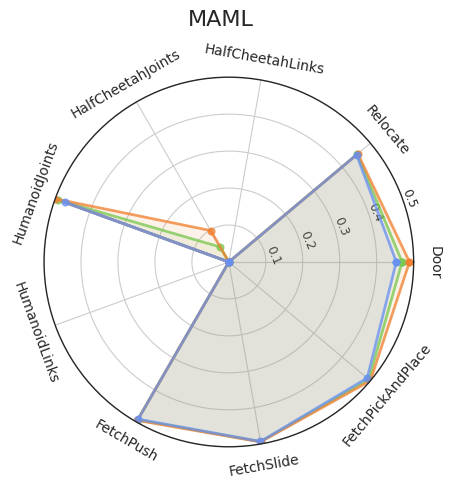}\label{fig:maml-raw} 
  }
  \subfigure[]{%
    \includegraphics[width=.3\columnwidth]{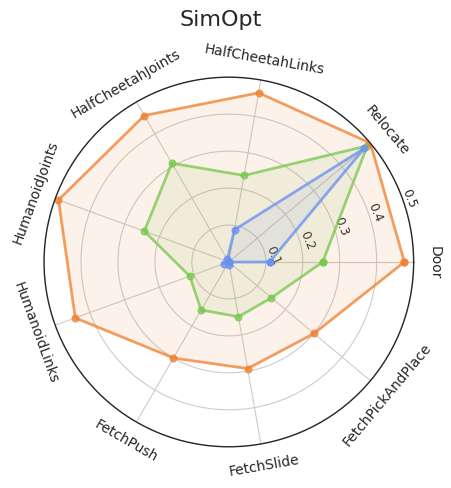}\label{fig:simopt-raw}
  } 
  
  \caption{All results for each estimator across each environment for experiments with direct regression without learning any policy; here, the maximum error (minimum accuracy) is shown in \textcolor{blue}{blue}, the mean accuracy shown in \textcolor{green}{green}, and the maximum accuracy shown in \textcolor{orange}{orange}.
  }
  \label{fig:rawresults}
\end{figure}

We see that while MAML performs strongly in lower dimensional settings, it struggles in higher-dimensional settings (Figure \ref{fig:maml-raw}); however, as the spread of the algorithm is remarkably low, it may be that with more extensive, particularly task-specific hyperparameter tuning, the algorithm can be made to work well in many parameter estimation settings. 

Both Active Domain Randomization (ADR) and SimOpt (Figures \ref{fig:adr-raw} and \ref{fig:simopt-raw} respecitively) do well on average, but have much larger spread - the variability between the individual particles is much higher here than with other algorithms tested.
Both algorithms are \textit{particle-based} estimators, rolling out and maintaining multiple estimates in a diversity-inducing (ADR) or evolutionary (SimOpt) manner. While some particles achieve high accuracy (the orange lines in each figure), the diversity-inducing behavior also works against both algorithms, leading to poor mean performance and a generally higher spread. 
SimOpt also tends to do poorly on lower-dimensional environments, likely due to the surjectiveness of the parameter estimation problem: many possible parameter settings, especially in low-dimensional problems, may lead to the same generated trajectory. ADR's explicit enforcement of diversity may allow it to refrain from latching onto these incorrect parameter estimates, whereas SimOpt inherently has no such protective mechanism.



\subsection{The Effect of Policy Learning}
\label{sec:policylearning}

The results in Section \ref{sec:dpe} were generated using static, held-out trajectories. However, a recent trend of work has been to do \textit{online} system-identification, especially in reinforcement learning contexts \cite{Rakelly:etal:2019, Nagabandi:etal:2020}. Here, rather than using expert data generated by a fully trained policy, we explore the effect of using an iteratively-learned policy to generate the trajectories for system identification.
In this section, we relax the assumption that the actions must be the same across trajectories, and rather use the policy to generate actions naturally (but control for starting states).

We use the \texttt{OurDDPG.py} variant from the open-source TD3 \cite{Fujimoto:etal:2018} repository, which is an improvement on the original Deep Deterministic Policy Gradient \cite{Lillicrap:etal:2015} algorithm. At the beginning of each episode, we randomize the simulator, and then roll out the policy in the generated environment. We use these same transitions to populate the replay buffer and train the policy. All trials are mean-averaged across three seeds, and each trial is used if and only if the policy trained converges to a ``environment-solving" solution.

\begin{figure}[h!]
  \centering
  \subfigure[]{%
    \includegraphics[width=.3\columnwidth]{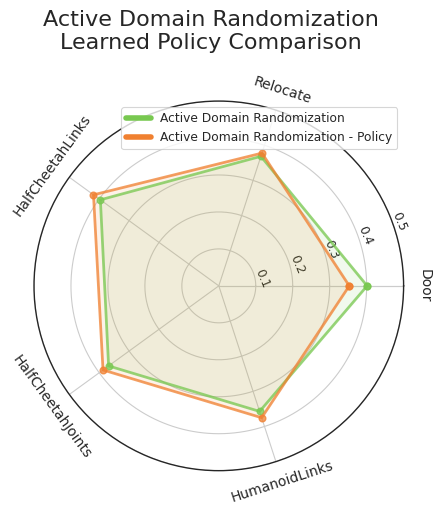}\label{fig:adr-policylearning}
  } ~
  \subfigure[]{%
    \includegraphics[width=.3\columnwidth]{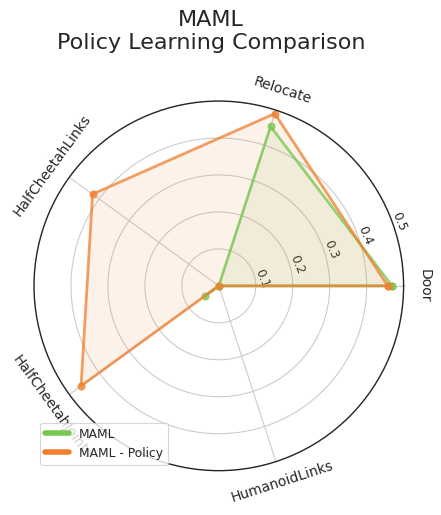}\label{fig:maml-policy} 
  } ~
  \subfigure[]{%
    \includegraphics[width=.3\columnwidth]{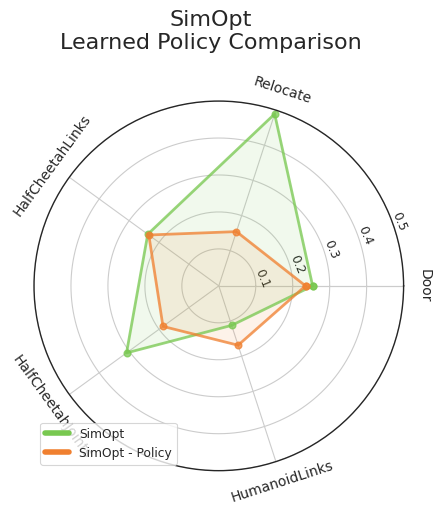}\label{fig:simopt-policylearning}
  }  
  
  \caption{Across five environments, we compare results for the effects of policy learning. The  \textcolor{green}{green} is the mean over trials when using demonstrations, while \textcolor{orange}{orange} shows the mean over trials when using the data generated by learning a policy. The results for linear regression and Bayesian Optimization can be found in Appendix \ref{app:fullresults}.}
  \label{fig:policylearning}
\end{figure}

As seen in Figure \ref{fig:policylearning}, we see that using a policy for trajectory generation stabilizes many of the algorithms --- the variance is significantly reduced as compared to Fig. \ref{fig:rawresults} --- despite the fact that the performance degrades in certain environments.

\subsection{The Effect of Learned Rewards}

In this section, we benchmark variants of \ac{ADR} and SimOpt using learned rewards (rather than the cost function being the MSE between states seen along the trajectory when executing the same actions). As the cost function is generated by training a discriminator, we roll out the actions from the same starting state, and train the discriminator to differentiate between the sources of trajectories. The benefit of using learned rewards is that starting states and actions no longer need to be held constant; however, this can make the parameter estimation problem more difficult.

\begin{figure}[h!]
  \centering
  \subfigure[]{%
    \includegraphics[width=.3\columnwidth]{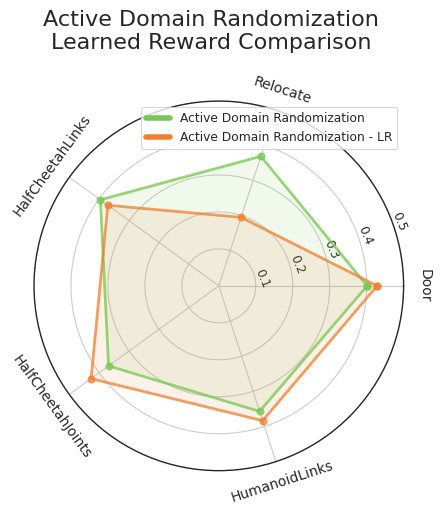}\label{fig:adr-learnedreward}
  } 
  \subfigure[]{%
    \includegraphics[width=.3\columnwidth]{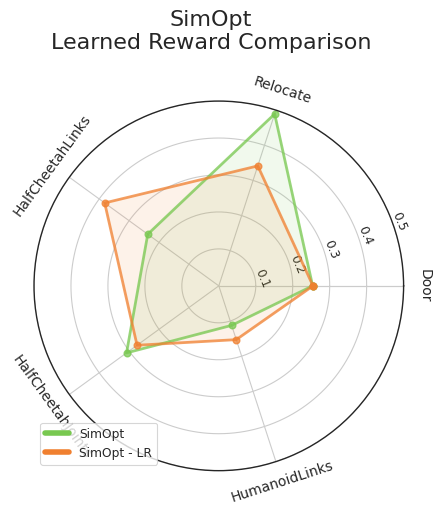}\label{fig:simopt-learnedreward}
  } 
  
  \caption{Across five environments, we compare results for the effects of learning rewards over using state-action differences. The  \textcolor{green}{green} is the mean over trials when using demonstrations, while \textcolor{orange}{orange} shows the mean over trials when using the data generated by learning a discriminator.}
  \label{fig:learnedreward}
\end{figure}

In our experiments, shown in Figure \ref{fig:learnedreward}, this difficulty seems to manifest itself in terms of \textbf{larger variance} between trials and generally \textbf{higher instability}. When using learned rewards, trends get less consistent. While both algorithms see less seeds converge, SimOpt seems to improve greatly from the use of a discriminator, whereas \ac{ADR} suffers from the use of learned rewards when compared to a cost function based on ground truth state differences.

\subsection{Generalization Experiments for Parameter Estimation via Domain Randomization}
\label{sec:domainrand}

\begin{figure}[h!]
\vspace{-5pt}
  \centering
  \includegraphics[width=1.0\columnwidth]{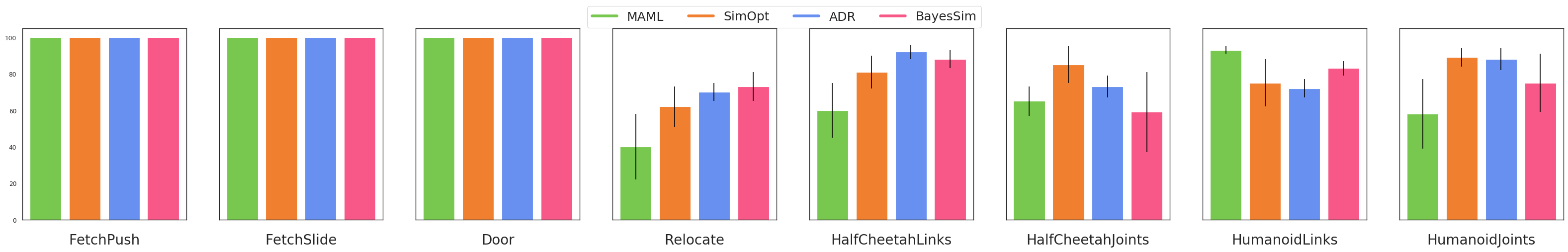}
  \caption{Across all environments, we compare the percentage  return of an agent trained in the proposed environment settings with respect to the return of an agent that is trained solely in the testing environment.}
  \vspace{-5pt}
  \label{fig:domainrand}

\end{figure}

Oftentimes, parameter estimation, system identification, or simulation calibration is the \textit{first} step of an engineering or research project. In this section, we benchmark each estimation algorithm in a \textit{zero-shot} transfer setting: using the estimated parameters, another agent is trained in the resulting parameter settings, and then tested on the held-out environment from which the reference (target) trajectories are generated. We compare the percentage return of an agent trained in the proposed environment settings normalised by the return of an agent that is trained solely in the target environment.


For MAML, as a point estimation algorithm, we train a reinforcement learning agent using the singular environment that is proposed after adapting to the reference trajectories. Since ADR and SimOpt are particle-based, we sample uniformly from the converged particles. In this section, we also compare against BayesSim, as comparing the agent's task performance is agnostic to the sampling strategy of environment generation.

We see in Figure \ref{fig:domainrand} that, generally, having a distribution to sample from improves test performance; however, our results seem to suggest that in these settings, recovery of the posterior is not critical for strong test time performance. Particles seem to suffice, and the two particle-based methods tested perform well on the high-dimensional, locomotion estimation tasks.

\section{Conclusion}

We introduce the \textbf{Si}mulation \textbf{P}arameter \textbf{E}stimation (\textbf{SIPE}) benchmark and provide extensive analysis of current methods in common and difficult parameter estimation tasks. We believe our codebase, library, and analysis will serve as a strong starting point for practitioners in the field. 



\clearpage
\acknowledgments{The authors would like to thank Yevgen Chebotar and Ian Abraham for their helpful comments and code implementation.}


\bibliography{paper}  

\newpage
\appendix

\section{Additional Results}
\label{app:additionalresults}

\begin{figure}
\centering
\begin{minipage}{.5\textwidth}
  \centering
  \includegraphics[width=1.0\columnwidth]{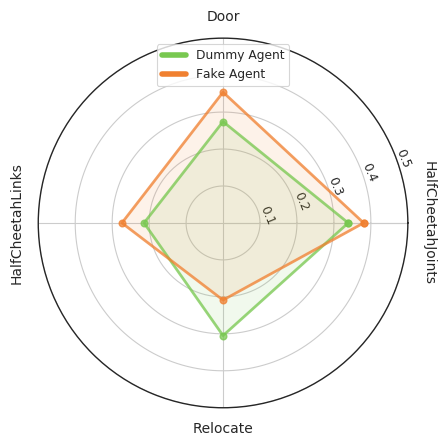}
    \caption{SiPE can also capture the spread of algorithms, enabling a better understanding of large-scale effects in high-dimensional settings, while still being able to understand the performance in one or two plots. Here, alongside \texttt{mean} averaging, we see that \texttt{max} (Figure \ref{fig:max-sipe}) and \texttt{min} (Figure \ref{fig:min-sipe}) provide additional insight to the two fake estimators shown here.}
  \label{fig:sipe-prototype}
\end{minipage}%
\begin{minipage}{.5\textwidth}
  \centering
  \begin{subfigure}
      \centering
      \includegraphics[width=0.5\columnwidth]{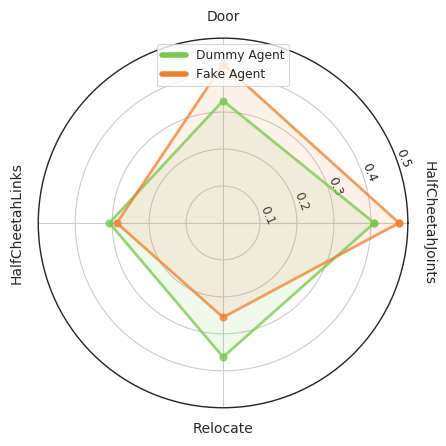}
      \caption{The \texttt{max} plotting capability of SiPE.}
      \label{fig:max-sipe}
    \end{subfigure}
    \begin{subfigure}
      \centering
      \includegraphics[width=0.5\columnwidth]{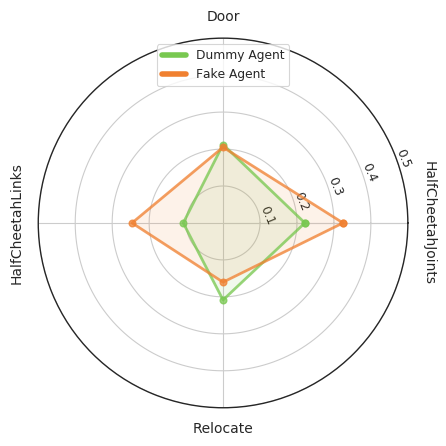}
      \caption{The \texttt{min} plotting capability of SiPE.}
      \label{fig:min-sipe}
    \end{subfigure}
\end{minipage}
\label{fig:sipe-extraprototypes}
\end{figure}

\subsection{Hyperparameters Discussion}
\label{app:hyperparameters}

As described in the main text, we used the validation set of trajectories from the SiPE benchmark to hyperparameter tune, choosing the best set of hyperparameters as measured by mean parameter accuracy on \textit{FetchPush-v1} and \textit{Relocate-v0}. We then held these parameters constant across the entire benchmark, and reported the performance both throughout the main text and Figure \ref{fig:bayesoptlr-results}. Within the following list, the \textbf{bolded} values are the chosen (benchmark) hyperparameters used in the study.

\begin{enumerate}
    \item \textbf{Regression}: For regression, we trained a single layer network (using a concatenated, per-timestep $s-a-s'$) to regress to the parameters directly. The hyperparameter search was over the batch size (\textbf{320}, 160, 80, 40) and learning rate (\textbf{0.001}, 0.01, 0.05, 0.005). We use the gradient descent version of linear regression, rather than using the closed form expression to solve for parameters.
    
    \item \textbf{Bayesian Optimization (BayesOpt)}: For Bayesian Optimization, we use the open-source \texttt{bayesopt} \cite{bayesoptgithub} package. We searched over the utility function (\textbf{upper-confidence bound}, expected improvement), $\kappa$ (\textbf{2.5}, 5, 10, 1, 0.5), and $\xi$ (\textbf{0}, 0.5, 1).
    
    \item \textbf{Model-Agnostic Meta-Learning (MAML)}: Using the regression task setup from \cite{maml}, we adapt the algorithm to the parameter estimation setting. For MAML, we use a two-layer hidden network (sizes searched over 40, \textbf{60}, and 100) and searched over a learning rate of 0.001, \textbf{0.005}, and 0.0001. We use the same learning rate for both the inner and outer loop. MAML was the only algorithm in the study implemented in Jax \cite{jax2018github}, while all other algorithms were implemented in Pytorch \cite{NEURIPS2019_9015}. 
    
    \item \textbf{SimOpt}: For SimOpt, our hyperparameter search was over the number of REPS updates (4, 8, \textbf{12}), the number of particles (4, \textbf{6}, 8), the mean initialization of the particles (mean sampled from U(0, 1) or \textbf{U(0, 0.5)}) and covariance initialization of the particles (0.05, 0.1, \textbf{0.2}). 
    
    \item \textbf{Active Domain Randomization}: For Active Domain Randomization, we searched over the number of particles (4, 6, 8, \textbf{10}) and discriminator learning rate for the learned reward experiments (searched over \textbf{0.001} and 0.005). We use the same policy network, temperature, discriminator learning rate and architecture as in \cite{adr}, and hold discriminator hyperparameters constant for the \textit{SimOpt \& Learned Reward} experiments. 
    
    \item \textbf{BayesSim}: We use the open-source implementation of BayesSim and do not modify any of the hyperparameters.
    
\end{enumerate}


\subsection{All Environment Results: Bayesian Optimization and Linear Regresssion}
\label{app:fullresults}
\begin{figure}[h!]
  \centering
  ~
  \subfigure[]{%
    \includegraphics[width=.3\columnwidth]{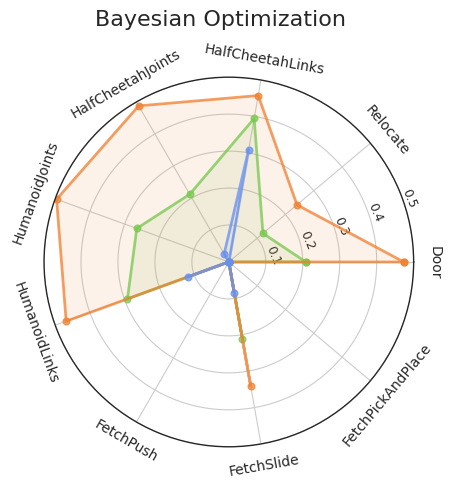}\label{fig:bayesopt-raw}
  } 
  ~ 
    \subfigure[]{%
    \includegraphics[width=.3\columnwidth]{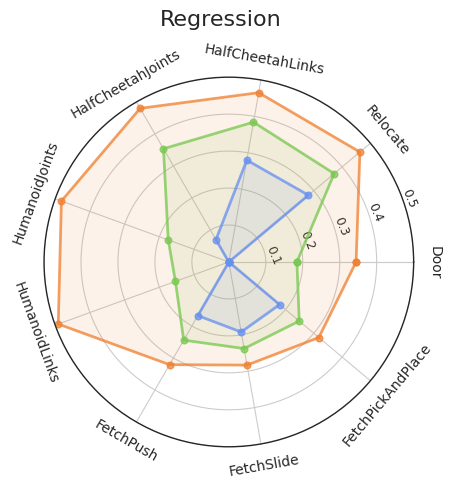}\label{fig:regression-raw}
  }
  \caption{All results for Linear Regression and Bayesian Optimization across each environment for experiments with direct regression without learning any policy; here, the maximum error (minimum accuracy) is shown in \textcolor{blue}{blue}, the mean accuracy shown in \textcolor{green}{green}, and the maximum accuracy shown in \textcolor{orange}{orange}}.
  \label{fig:bayesoptlr-results}
\end{figure}

For completeness, we provide results analogous to those seen in Section \ref{sec:dpe} for Linear Regression and Bayesian Optimization. As seen in Figures \ref{fig:bayesopt-raw} and \ref{fig:regression-raw}, these simplest approaches seems to do well on high dimensional environments (Bayesian Optimization) or well generally (Regression). While a high spread, these results provide a practical suggestion: sometimes, simple methods can work just as well, especially with further tuning to a specific problem of interest. 

\subsection{Policy Learning and Learned Rewards}

When we combine learning a policy and using learned rewards, we see that SimOpt varies highly in performance, greatly improving on the high-dimensional estimation task of \texttt{HumanoidLinks}. 

\begin{figure}[h!]
  \centering
  \subfigure[]{%
    \includegraphics[width=.3\columnwidth]{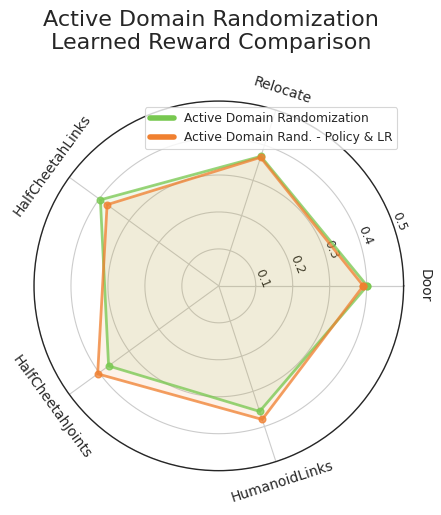}\label{fig:adr-policylearnedreward}
  } 
  \subfigure[]{%
    \includegraphics[width=.3\columnwidth]{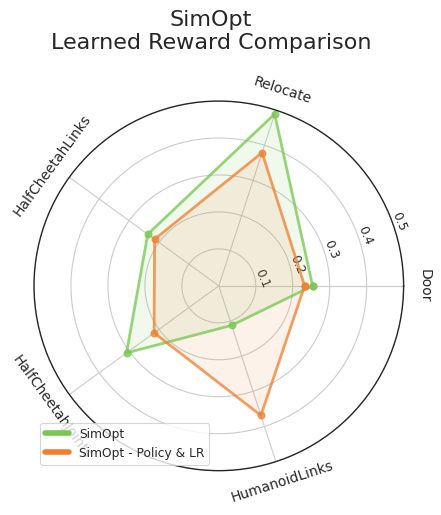}\label{fig:simopt-policylearnedreward}
  } 
  \caption{Across five environments, we compare results for the effects of learning rewards \textbf{and} using trajectories from a policy over using state-action differences. The  \textcolor{green}{green} is the mean over trials when using demonstrations, while \textcolor{orange}{orange} shows the mean over trials when using the data generated by learning a discriminator.}
  \label{fig:policyandlr}
\end{figure}

For ADR, we see that learning rewards alongside a policy does not provide extra benefit over learning solely a policy. The two in conjunction seem to destabilize approaches (i.e fewer seeds reach the \textit{Convergence} score), which we attribute to the learned rewards.

\subsection{Fine-tune or Adapt?}

\begin{figure}
\centering
\begin{minipage}{.5\textwidth}
  \centering
  \includegraphics[width=0.8\columnwidth]{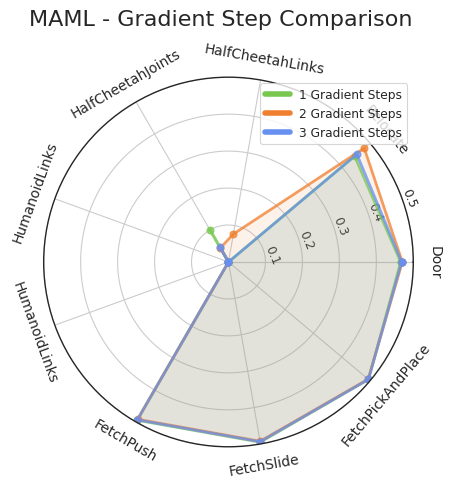}
  \caption{Across all environments, we compare results for the effects of gradient steps when using MAML.}
  \label{fig:maml-gradientsteps}
\end{minipage}%
\begin{minipage}{.5\textwidth}
  
\end{minipage}
\end{figure}

When we test MAML with varying numbers of \textit{test-time} gradient steps, we see that while the gradient steps hardly affect performance, MAML exhibits \textit{all-or-nothing} behavior. As noted in the experimental setup, our hyperparameters are tuned on two environments which receive almost 100\% performance. However, as shown in Figure \ref{fig:maml-gradientsteps}, MAML seems to have issues with the higher-dimensional parameter estimation tasks. This may be a result of fundamental issues with the algorithm, or a need for more extensive tuning of hyperparameters. 

\end{document}